\documentclass[11pt]{article}
\usepackage[preprint]{acl}

\usepackage{times}
\usepackage{latexsym}
\usepackage[T1]{fontenc}
\usepackage[utf8]{inputenc}
\usepackage{microtype}
\usepackage{inconsolata}
\usepackage{graphicx}

\usepackage{amsmath,amssymb}
\newcommand{\xmark}{\ensuremath{\times}}
\newcommand{\cmark}{\ensuremath{\checkmark}}
\usepackage{booktabs}
\usepackage{multirow}
\usepackage{placeins}
\usepackage{algorithm,algpseudocode}
\usepackage{xcolor}
\usepackage{listings}
\usepackage{enumitem}

\lstdefinelanguage{json}{
  morestring=[b]",
  morecomment=[l]{//},
  showstringspaces=false,
  literate=
    *{0}{{{\color{purple}0}}}{1}
     {1}{{{\color{purple}1}}}{1}
     {2}{{{\color{purple}2}}}{1}
     {3}{{{\color{purple}3}}}{1}
     {4}{{{\color{purple}4}}}{1}
     {5}{{{\color{purple}5}}}{1}
     {6}{{{\color{purple}6}}}{1}
     {7}{{{\color{purple}7}}}{1}
     {8}{{{\color{purple}8}}}{1}
     {9}{{{\color{purple}9}}}{1}
     {:}{{{\color{black}{:}}}}{1}
     {,}{{{\color{black}{,}}}}{1}
     {\{}{{{\color{black}{\{}}}}{1}
     {\}}{{{\color{black}{\}}}}}{1}
     {[}{{{\color{black}{[}}}}{1}
     {]}{{{\color{black}{]}}}}{1},
}
\lstdefinestyle{cica-json}{
  language=json,
  basicstyle=\ttfamily\small,
  stringstyle=\color{teal},
  commentstyle=\color{gray}\itshape,
  keywordstyle=\color{blue}\bfseries,
  morekeywords={true,false,null},
  showstringspaces=false,
  breaklines=true,
  breakatwhitespace=true,
  columns=fullflexible,
  keepspaces=true,
  frame=none,
  tabsize=2,
}

\title{Diagnosis Is Not Prescription: Linguistic Co-Adaptation Explains Patching Hazards in LLM Pipelines}

\author{
  Jeonghun Yoon\textsuperscript{1,2}\thanks{\,Corresponding author.} \quad Dongchan Kim\textsuperscript{3} \\
  \textsuperscript{1}KAIST (Korea Advanced Institute of Science and Technology), Republic of Korea \\
  \textsuperscript{2}NAVER Corp., Seongnam, Republic of Korea \quad
  \textsuperscript{3}NAVER Corp., Bellevue, WA, USA \\
  \texttt{jeonghun.yoon@kaist.ac.kr} \quad \texttt{homer.yoon@navercorp.com} \\
  \texttt{dongchan.usa@gmail.com} \\
}

\begin{document}
\maketitle

\begin{abstract}
When a multi-module LLM agent fails, the module most responsible
for the failure is not necessarily the best place to intervene.
We demonstrate this \textbf{Diagnostic Paradox} empirically: causal
analysis consistently identifies the \emph{routing module} --- which
selects which tool to call next --- as the primary bottleneck across
three independent agent families. Yet injecting prompt-level
correction examples into this module consistently degrades
performance, sometimes severely. Patching an upstream
\emph{query-rewriting module} instead reliably improves outcomes.
The effect holds with statistical significance on two agent families
and directional consistency on a third; alternative repair strategies
at the routing module (instruction rewriting, model upgrade) are
neutral, confirming that the harm is specific to correction-injection
patching.

We explain this asymmetry through the \textbf{Linguistic Contract}
hypothesis: each downstream module implicitly adapts to its
upstream's characteristic error distribution, so correcting the
bottleneck breaks this implicit alignment in a way that upstream
corrections do not. We operationalize this via a per-agent
co-adaptation measure, derived from diagnosis alone, and show it
is consistently associated with patching harm across agent families:
higher co-adaptation co-occurs with harm, lower with safety. This
trend holds across all three agent families, providing preliminary
support for the hypothesis beyond a single-agent observation.
\end{abstract}

\begin{figure}[t]
\centering
\includegraphics[width=\linewidth]{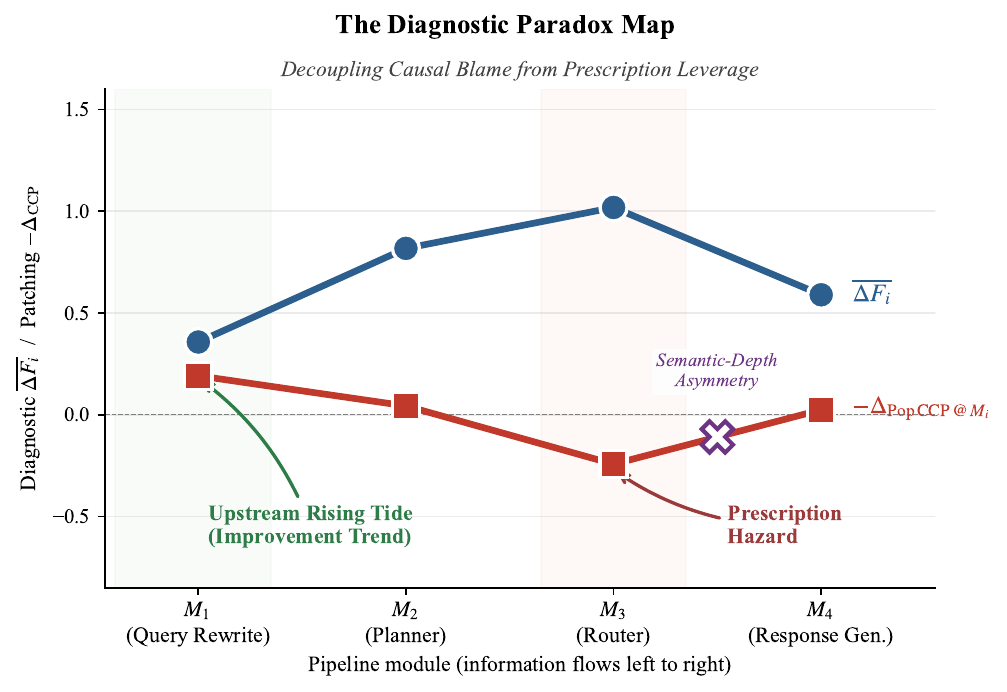}
\caption{\textbf{Diagnosis Is Not Prescription.}
Blue ($\overline{\Delta F_i}$): causal contribution of each
module on the diagnosis set---the routing module ($M_3$) peaks.
Red ($-\Delta_{\mathrm{CCP}\,@\,M_i}$): performance change from
patching each module---negative at $M_3$ (harm), positive at
$M_1$ (improvement). The primary statistically robust finding
is the $M_1$-vs.-$M_3$ contrast ($p{<}0.01$, Holm-corrected);
$M_1$'s standalone improvement does not survive correction
($p{=}0.086$).}
\label{fig:diagnostic-paradox-map}
\end{figure}

\section{Introduction}
\label{sec:intro}

Modern LLM agents chain multiple NLP modules in sequence: a
query rewriter clarifies the user's intent, a planner decomposes
it into steps, a router selects the appropriate tool, and a
response generator produces the final
reply~\citep{khattab2024dspy,yao2023react,shinn2023reflexion,yao2024tau}.
Each module is an LLM call, and their sequential arrangement
introduces a critical challenge: when the agent fails, the
failure could originate anywhere in the chain---and the optimal
place to intervene may not be where the failure originates.

Existing approaches---end-to-end
optimizers~\citep{yuksekgonul2024textgrad,cheng2024trace} and
process reward models~\citep{lightman2023verify}---treat diagnosis
and prescription as equivalent. We show they are not.

We study this across three independent base agents using causal
intervention analysis and discover the \textbf{Diagnostic Paradox}
(Figure~\ref{fig:diagnostic-paradox-map}): the Router module, which causes
the most failures, is the worst place to apply a prompt-level
correction. Patching it consistently degrades performance, while
patching the upstream Query Rewriter helps. The effect is
robust---statistically significant on two agents and directionally
consistent on a third.

We explain this through the \textbf{Linguistic Contract}
hypothesis: downstream NLP modules implicitly adapt to their
upstream's output distribution, including its characteristic
errors. Correcting the diagnosed bottleneck changes this
distribution in a way downstream modules cannot recover from,
while upstream corrections propagate through co-adapted chains
more gracefully. We operationalize Contract strength via a
measurable per-task quantity and show it exhibits a consistent
trend with patching Hazard across all three agent families.

We study a four-module pipeline on $\tau$-bench
retail~\citep{yao2024tau}: $M_1$ (query rewrite) $\to$ $M_2$
(planner) $\to$ $M_3$ (router) $\to$ $M_4$ (response generator).
Formal definitions are in App.~\ref{app:prelim}.

\noindent\textbf{Contributions:}
\begin{itemize}[leftmargin=*,nosep,topsep=2pt,itemsep=1pt]
\item \textbf{Diagnostic Paradox:} through causal intervention
  experiments on a multi-module LLM pipeline, we find the
  diagnosed bottleneck is the worst correction-patch target;
  upstream modules are more effective. Demonstrated across three
  agent families (two with significance, one directional).
\item \textbf{Linguistic Contract:} downstream modules co-adapt
  to upstream error distributions; a diagnostic measure of this
  co-adaptation shows a consistent trend with patching Hazard
  across all three agent families.
\end{itemize}

\section{Related Work}
\label{sec:related}

Pipeline debugging in NLP has historically relied on
component-wise analysis and error propagation
studies~\citep{khattab2024dspy,shinn2023reflexion}.
End-to-end optimizers such as
TextGrad~\citep{yuksekgonul2024textgrad} and
TRACE~\citep{cheng2024trace} improve overall pipeline performance
but do not separate \emph{where blame lies} from
\emph{where to intervene}. Process reward
models~\citep{lightman2023verify} localize step-level errors but
require costly human annotation. Our work fills this gap with a
causal framework operating at the LLM-call level.

Causal mediation analysis~\citep{pearl2009causality,pearl2014interpretation}
has been applied inside single models at the neuron and
attention-head level~\citep{vig2020causal,stolfo-etal-2023-causal}
to understand how linguistic information flows. We apply the same
principle one level up---across full LLM module calls in a
pipeline---and discover that the module-level co-adaptation
dynamics produce counter-intuitive repair properties not visible
at the token level. CCP differs from iterative
self-refinement~\citep{madaan2023selfrefine} and prompt
optimization~\citep{yang2024opro,zhou2023ape} by reusing
causally curated correction triples with no additional inference.
The cross-family agent/oracle/judge design prevents LLM-judge
self-correlation~\citep{liu2023g}.

\section{Method: CICA and the Linguistic Contract}
\label{sec:method}

CICA (\emph{Causal Intervention-based Analysis}) comprises four
layers: failure index $F$ (\S\ref{ssec:fi}), causal contribution
$\Delta F_i$ (\S\ref{ssec:deltaF}), per-task fates via
$\mathrm{NIE}_i$ (\S\ref{ssec:nie-method}), and CCP with
module-normalized routing (\S\ref{ssec:ccp}),
connected by the Linguistic Contract (\S\ref{ssec:contract}).

\subsection{Failure Index $F$: System-Level Symptom}
\label{ssec:fi}

An LLM judge scores per-module severities $\mathrm{sev}_i \in [0,1)$;
the failure index aggregates them:
\begin{equation}
F(E) = -\sum_{i=1}^{k} \log\!\bigl(1-\mathrm{sev}_i\bigr).
\label{eq:fi}
\end{equation}
Every prescription configuration aims to reduce $\overline{F}$
on $\mathcal{D}_{\mathrm{presc}}$. Negative $\Delta{=}F_{\mathrm{patched}}-F_{\mathrm{base}}$ indicates improvement.
Intuitively, lower $F$ means the agent behaved better: when the
agent selects the correct first action, $\bar{F}{=}1.87$; when it
does not, $\bar{F}{=}4.69$ (gap $2.81$).\footnote{Pearson
$r{=}{-}0.90$ ($p{<}10^{-40}$) and Spearman $\rho{=}{-}0.83$
between per-task $F$ and first-action tool-name match on the
prescription baseline ($n_p{=}111$); App.~\ref{app:f-validity}.}
$F$ thus reflects behavioral outcomes, not merely
internal judge scores.

\subsection{Causal Contribution $\Delta F_i$: Locating the Failure Origin}
\label{ssec:deltaF}

We replace $M_i$'s output with oracle $S^*_i$ and re-execute:
\begin{equation}
\Delta F_i = F(E) - F\bigl(E \mid \mathrm{do}(M_i = S^*_i)\bigr).
\label{eq:deltaF}
\end{equation}
Larger $\Delta F_i$ means $M_i$ is more causally responsible. The
population diagnostic target $i^\star_{\mathrm{pop}} = \arg\max_i
\overline{\Delta F_i}$ (where the bar denotes the mean of $\Delta F_i(c)$
over $c \in \mathcal{D}_{\mathrm{diag}}$)
drives the matched Population-level CCP (Pop CCP) measurement that
exposes the Diagnostic Paradox (\S\ref{ssec:paradox}).

\subsection{Per-task Fates: Natural Indirect Effect $\mathrm{NIE}_i$}
\label{ssec:nie-method}

$\mathrm{NIE}_i$ isolates the upstream-cleanup effect flowing
\emph{through} $M_i$: two worlds share oracle-clean upstream but
differ in whether $M_i$ re-executes or is frozen at baseline
(App.~\ref{app:nie}). With threshold $\tau{=}0.05$, each
(module, task) pair is a \textbf{compensator}
($\mathrm{NIE}_i{<}{-}\tau$: downstream absorbs upstream noise),
\textbf{amplifier} ($\mathrm{NIE}_i{>}{+}\tau$: it amplifies), or
\textbf{propagator} (neutral). The fraction of compensator tasks
at $M_3$ is our proxy for \emph{Contract strength}
(\S\ref{ssec:contract}).

\subsection{Non-destructive Repair: CCP}
\label{ssec:ccp}

CCP appends $k{=}5$ severity-ranked correction triples
$(\text{input}, \text{wrong}, \text{oracle})$ from
$\mathcal{D}_{\mathrm{diag}}$ to the target module's prompt, with
frozen instructions and output schema. Downstream modules
thus see outputs from the same distributional family as baseline---
distinguishing CCP from rewriting, which alters instructions and
triggers a different downstream cascade. Details in
App.~\ref{app:ccp-pool}. An adaptive $z$-score routing variant
normalizes per-module NIE heterogeneity (App.~\ref{app:a6-proc}).

\subsection{The Linguistic Contract Hypothesis}
\label{ssec:contract}

We name the theoretical structure that connects all four CICA
layers. Each downstream module implicitly adapts to its upstream's
output distribution---including its characteristic errors---developing
compensation behaviors we call the \textbf{Linguistic Contract}.
This is not a formal agreement but an emergent property: modules
trained or prompted on a given upstream become sensitive to its
specific output patterns. When a prescription corrects the bottleneck,
this alignment breaks: compensation fails, previously masked errors
surface, and the system regresses (the Prescription Hazard).

The decisive distinction between patching $M_3$ vs.\ $M_1$ is
direction: $M_3$ CCP perturbs an executable-semantic layer
(tool-name routing) that downstream modules cannot absorb; $M_1$
CCP perturbs a surface-linguistic layer that downstream modules
naturally compensate for (\S\ref{ssec:mechanism}).

We distinguish two forms: \emph{Relative} ($M_3$ is the worst patch
target) and \emph{Absolute} ($\Delta(M_3){>}0$, only when Contract
strength is high). We use the $M_3$ compensator rate as a preliminary
proxy for Contract strength; empirical values and the resulting
Hazard trend are reported in \S\ref{ssec:paradox} and
App.~\ref{app:cross-agent}.

\section{Experiments}
\label{sec:experiments}

\begin{table}[t]
\footnotesize\centering
\setlength{\tabcolsep}{3pt}
\renewcommand{\arraystretch}{1.1}
\begin{tabular}{@{}lcccc@{}}
\toprule
\textbf{Mod.}
 & \textbf{$\overline{\Delta F_i}$}
 & \textbf{$\overline{\mathrm{NIE}_i}$}
 & \textbf{$\sigma_{\mathrm{NIE}}$}
 & \textbf{Amp/Prop/Comp} \\
\midrule
$M_2$ (Pl.)         & $+$0.817 & $-$0.693 & 0.638 & 40\,/\,\phantom{0}6\,/\,454 \\
$M_3$ (R.)$^\dagger$  & $+$1.018 & $-$1.014 & 0.572 & \phantom{0}2\,/\,\phantom{0}7\,/\,491 \\
$M_4$ (RG)          & $+$0.589 & $-$0.589 & 0.221 & \phantom{0}0\,/\,\phantom{0}8\,/\,492 \\
\bottomrule
\end{tabular}
\caption{Per-module diagnostic stats ($n_d{=}500$,
$\mathcal{D}_{\mathrm{diag}}$, gpt-4o-mini). $^\dagger$$M_3$ is
$i^\star_{\mathrm{pop}}$. Right half: per-task fate counts
($|\mathrm{NIE}_i|{>}0.05$). $M_1$ excluded
($\overline{\Delta F_1}{=}0.357$, enters via
\S\ref{ssec:paradox}). Full mediation triple in
App.~\ref{app:nie}.}
\label{tab:e1e2}
\end{table}

Two questions: \textbf{RQ1}~is the diagnosed module ($M_3$) also
the worst CCP target (\emph{the Diagnostic Paradox})? \textbf{RQ2}~what
mechanism explains this gap (\emph{mechanistic account, exploratory})?

\subsection{Setup}
\label{ssec:setup}

\paragraph{Benchmark and splits.}
$\tau$-bench retail~\citep{yao2024tau}: train split is the
diagnosis set $\mathcal{D}_{\mathrm{diag}}$ ($n_d{=}500$),
test split is the prescription set $\mathcal{D}_{\mathrm{presc}}$
($n_p{=}111$ valid; disjoint). All diagnosis-derived artifacts
($i^\star_{\mathrm{pop}}$, CCP pool, fates, $(\mu_i,\sigma_i)$)
come from $\mathcal{D}_{\mathrm{diag}}$ only.
$\tau$-bench airline~\citep{yao2024tau} is used for diagnostic
generalization only ($n{=}37$; no prescription split).

\paragraph{Agent families.}
We use three base agents from distinct model families:
gpt-4o-mini (OpenAI), Llama~4 Scout (Meta), and Qwen3-32b
(Alibaba). Each is paired with the same
oracle (Claude~Sonnet~4.6, Anthropic) and LLM judge
(Gemini~2.5~Flash, Google); this cross-family triple prevents
judge--oracle self-correlation~\citep{liu2023g}. Full
prescription experiments are run on gpt-4o-mini and Qwen3-32b
on $\mathcal{D}_{\mathrm{presc}}$; Llama~4 Scout provides
directional prescription replication. Diagnostic sweeps cover
all three agents. See App.~\ref{app:models} for model details.

\subsection{Configurations and Statistical Analysis}
\label{ssec:configs}

The main configuration sweep applies Pop CCP at each of
$M_1$--$M_4$ with $k{=}5$ severity-ranked correction triples,
plus a $z$-score adaptive routing variant and an oracle
upper-bound injection. Ablations test alternate repair strategies
at $M_3$ (full list in App.~\ref{app:configurations}).
Significance is assessed via paired Wilcoxon tests with
Holm--Bonferroni correction over four pre-specified comparisons
(Table~\ref{tab:pvalues}); effect size is paired Cohen's $d_z$.
The Qwen3-32b prescription pilot uses the same correction pool
constructed from gpt-4o-mini $\mathcal{D}_{\mathrm{diag}}$
(App.~\ref{app:cross-agent}).

\subsection{The Diagnostic Paradox}
\label{ssec:paradox}

\paragraph{Diagnosis: identifying the bottleneck.}
CICA identifies the router ($M_3$) as the population diagnostic
target with the largest causal contribution
($\overline{\Delta F_3}{=}1.018$; Table~\ref{tab:e1e2}), ahead
of the planner ($0.817$), response generator ($0.589$), and
query rewriter ($0.357$). This result generalizes across both domains and agent families.
On $\tau$-bench airline, $M_3$ again ranks first
($\overline{\Delta F_3}{=}1.69$; App.~\ref{app:airline-e2}),
as it does when Llama~4 Scout ($1.564$) and Qwen3-32b ($1.148$)
replace gpt-4o-mini as the base agent on retail
(App.~\ref{app:cross-agent}).

\paragraph{Prescription: the diagnosed bottleneck is the worst patch target.}
Table~\ref{tab:prescription-summary} summarises the key prescription
results across all three agents; full per-configuration details are
in App.~\ref{app:cross-agent}. The $M_3$-worst, $M_1$-best ordering
holds consistently: $M_3$ CCP produces degradation on the two
high-compensator agents (absolute Hazard), while $M_1$ CCP yields
the largest improvement of any agent tested. Baseline-vs.-$M_1$
alone does not survive Holm correction on any agent
($p{\geq}0.086$); the Paradox is established by the
$M_1$-vs.-$M_3$ contrast.

\begin{table}[t]
\footnotesize\centering
\setlength{\tabcolsep}{4pt}
\renewcommand{\arraystretch}{1.1}
\begin{tabular}{@{}l r r r@{}}
\toprule
\textbf{Agent} & \textbf{@ $M_3$} & \textbf{@ $M_1$} & \textbf{contrast} \\
\midrule
gpt-4o-mini   & $+0.243^{\ast}$         & $-0.191$ & $+0.435^{\ast\ast}$ \\
Qwen3-32b     & $+0.683^{\ast\ast\ast}$ & $-0.659$ & $+1.343^{\ast\ast\ast}$ \\
Llama~4       & $-0.013$                & $-0.402$ & $+0.389$ \\
\bottomrule
\end{tabular}
\caption{Prescription $\Delta$ for Pop CCP. $M_3$-worst, $M_1$-best.
$^{\ast}$Holm $p{<}0.05$; $^{\ast\ast}p{<}0.01$; $^{\ast\ast\ast}$raw $p{<}0.001$;
no mark = not significant after Holm. Full results in App.~\ref{app:cross-agent}.}
\label{tab:prescription-summary}
\end{table}

\paragraph{Contract trend: consistent across agent families.}
The Linguistic Contract hypothesis suggests that agents with
heavy downstream co-adaptation should exhibit CCP-induced
degradation, while agents without co-adaptation should not.
We measured compensator rates on $\mathcal{D}_{\mathrm{diag}}$
before any prescription experiment, then observed Hazard on
$\mathcal{D}_{\mathrm{presc}}$: the two high-rate agents
(gpt-4o-mini $98.2\%$, Qwen3-32b $96.0\%$) both show absolute
Hazard, while Llama~4 Scout ($0\%$) does not. This consistent
trend across three agent families---with measurements derived
from diagnosis before any prescription run---supports the
Contract hypothesis without establishing it causally.
Cross-judge $M_3$ agreement: $\alpha{=}0.128$ (ceiling-compressed;
binary 80\%, direction 84\%; App.~\ref{app:judge-agreement}).

\paragraph{CCP-specificity.}
The Hazard is specific to CCP: alternative repair strategies
at $M_3$ (instruction rewriting, model upgrade, task-type
routing) all produce $\Delta{\approx}0$
(App.~\ref{app:configurations}). This scoping rules out the
interpretation that $M_3$ is simply a fragile module---rather,
the Hazard arises from CCP's particular mechanism of injecting
correction examples that shift $M_3$'s output distribution.
As an upper bound, oracle injection at $M_3$ (replacing $M_3$'s
output with a cached gold action) yields the strongest improvement
on gpt-4o-mini ($\Delta{=}{-}2.007$) and Llama~4 Scout
($\Delta{=}{-}1.795$; App.~\ref{app:cross-agent}); this
configuration was not tested on Qwen3-32b. The result confirms
that $M_3$ is improvable---just not via CCP.

\subsection{Mechanistic Account (Exploratory)}
\label{ssec:mechanism}

Cascade analysis (App.~\ref{app:cascade}) gives a candidate
distinction: \emph{magnitude of downstream shift is
patch-location-invariant}. $M_4$ cosine to baseline is
$0.837/0.906$ (BoW / sentence-encoder all-MiniLM-L6-v2) under
Pop CCP @ $M_3$ vs.\ $0.839/0.906$ under Pop CCP @ $M_1$
(paired Wilcoxon $p{=}0.76$, $n{=}111$), yet $\Delta F$ differs
by $0.434$ ($+0.243$ vs.\ $-0.191$) in opposite directions.
The two locations diverge at $M_3$ itself
(sentence-encoder $0.920$ vs.\ $0.959$, $p{=}0.006$), suggesting
the divergence localizes to the executable-semantic layer
($9.9\%$ tool-name disagreement at $M_3$ vs.\ mostly
surface-text shift at $M_1$). We frame this as suggestive
rather than confirmed: the directional account is inferred from
the joint pattern of comparable cosine and opposite outcome,
and structural confounds ($M_3$ categorical vs.\ $M_1$
continuous outputs) remain (\S\ref{sec:limitations} item~6).
A direct geometric projection onto an oracle-aligned axis would
constitute cleaner evidence; App.~\ref{app:ccp-examples}.

\section{Conclusion}
\label{sec:conclusion}

We demonstrated the Diagnostic Paradox: in a
multi-module LLM pipeline, the module with the highest causal
blame is the worst target for prompt-level correction patching,
while upstream modules are more effective. In our pipeline, this
manifests as the router ($M_3$, diagnosed bottleneck) being
consistently outperformed by patching the query rewriter
($M_1$, upstream) across three independent agent families.
The Linguistic Contract hypothesis---downstream modules
co-adapt to upstream noise---shows a consistent trend with
patching Hazard across all three agent families.

\section{Limitations}
\label{sec:limitations}

\begin{enumerate}\itemsep=0pt
\item \textbf{Scope of evidence.} The diagnostic finding ($M_3$
  as bottleneck) generalizes across three agent families and two
  domains (retail and airline). Prescription evidence, however,
  remains limited to $\tau$-bench retail with one fixed pipeline
  topology; whether the patching Hazard persists in other task
  domains or architectures requires further study.
\item \textbf{Single-turn first-action protocol.} Causal mediation
  requires per-stage oracle pairing, well-defined only at one
  decision point; multi-turn extension is deferred. We expect the
  Paradox direction to persist under multi-turn evaluation, but
  this is plausibility, not evidence.
\item \textbf{Oracle/judge design.} The cross-family oracle/judge
  triple mitigates self-correlation; an inter-judge audit confirms
  substantial agreement (App.~\ref{app:judge-agreement}).
  Human evaluation remains future work.
\item \textbf{Fate threshold $\tau$.} The Paradox finding is robust
  across $\tau\in\{0.01,0.05,0.10,0.20\}$
  (App.~\ref{app:tau-sensitivity}).
\item \textbf{Directional account is indirect.} The direction
  hypothesis is inferred from comparable $M_4$ cosine yet opposite
  $\Delta F$ outcomes. A direct geometric projection onto an
  oracle-aligned axis would constitute cleaner evidence.
\item \textbf{Contract proxy is NIE-derived.} Compensator rate
  is derived from the same $\mathrm{NIE}_i$ used in diagnosis; an
  independent co-adaptation measure is needed to establish a causal
  mechanism rather than a correlational indicator.
\end{enumerate}

\bibliography{references}

\begin{thebibliography}{15}
\providecommand{\natexlab}[1]{#1}

\bibitem[{Cheng et~al.(2024)Cheng, Nie, and Swaminathan}]{cheng2024trace}
Ching-An Cheng, Allen Nie, and Adith Swaminathan. 2024.
\newblock {TRACE} is the next {AutoDiff}: Training compute-optimal language
  model agents.
\newblock \emph{arXiv preprint arXiv:2406.16218}.

\bibitem[{Khattab et~al.(2024)Khattab, Singhvi, Maheshwari, Zhang, Santhanam,
  Vardhamanan, Haq, Sharma, Joshi, Moazam, Miller, Zaharia, and
  Potts}]{khattab2024dspy}
Omar Khattab, Arnav Singhvi, Paridhi Maheshwari, Zhiyuan Zhang, Keshav
  Santhanam, Sri Vardhamanan, Saiful Haq, Ashutosh Sharma, Thomas~T. Joshi,
  Hanna Moazam, Heather Miller, Matei Zaharia, and Christopher Potts. 2024.
\newblock {DSPy}: Compiling declarative language model calls into
  self-improving pipelines.
\newblock In \emph{International Conference on Learning Representations
  (ICLR)}.

\bibitem[{Lightman et~al.(2023)Lightman, Kosaraju, Burda, Edwards, Baker, Lee,
  Leike, Schulman, Sutskever, and Cobbe}]{lightman2023verify}
Hunter Lightman, Vineet Kosaraju, Yura Burda, Harri Edwards, Bowen Baker, Teddy
  Lee, Jan Leike, John Schulman, Ilya Sutskever, and Karl Cobbe. 2023.
\newblock Let's verify step by step.
\newblock \emph{arXiv preprint arXiv:2305.20050}.

\bibitem[{Liu et~al.(2023)Liu, Iter, Xu, Wang, Xu, and Zhu}]{liu2023g}
Yang Liu, Dan Iter, Yichong Xu, Shuohang Wang, Ruochen Xu, and Chenguang Zhu.
  2023.
\newblock {G-Eval}: {NLG} evaluation using {GPT-4} with better human alignment.
\newblock In \emph{Proceedings of the 2023 Conference on Empirical Methods in
  Natural Language Processing (EMNLP)}.

\bibitem[{Madaan et~al.(2023)Madaan, Tandon, Gupta, Hallinan, Gao, Wiegreffe,
  Alon, Dziri, Prabhumoye, Yang, Gupta, Majumder, Hermann, Welleck,
  Yazdanbakhsh, and Clark}]{madaan2023selfrefine}
Aman Madaan, Niket Tandon, Prakhar Gupta, Skyler Hallinan, Luyu Gao, Sarah
  Wiegreffe, Uri Alon, Nouha Dziri, Shrimai Prabhumoye, Yiming Yang, Shashank
  Gupta, Bodhisattwa~Prasad Majumder, Katherine Hermann, Sean Welleck, Amir
  Yazdanbakhsh, and Peter Clark. 2023.
\newblock Self-refine: Iterative refinement with self-feedback.
\newblock In \emph{Advances in Neural Information Processing Systems
  (NeurIPS)}.

\bibitem[{Pearl(2009)}]{pearl2009causality}
Judea Pearl. 2009.
\newblock \emph{Causality: Models, Reasoning and Inference}, 2nd edition.
\newblock Cambridge University Press.

\bibitem[{Pearl(2014)}]{pearl2014interpretation}
Judea Pearl. 2014.
\newblock Interpretation and identification of causal mediation.
\newblock \emph{Psychological Methods}, 19(4):459--481.

\bibitem[{Shinn et~al.(2023)Shinn, Cassano, Berman, Gopinath, Narasimhan, and
  Yao}]{shinn2023reflexion}
Noah Shinn, Federico Cassano, Edward Berman, Ashwin Gopinath, Karthik
  Narasimhan, and Shunyu Yao. 2023.
\newblock Reflexion: Language agents with verbal reinforcement learning.
\newblock In \emph{Advances in Neural Information Processing Systems
  (NeurIPS)}.

\bibitem[{Stolfo et~al.(2023)Stolfo, Belinkov, and
  Sachan}]{stolfo-etal-2023-causal}
Alessandro Stolfo, Yonatan Belinkov, and Mrinmaya Sachan. 2023.
\newblock A mechanistic interpretation of arithmetic reasoning in language
  models using causal mediation analysis.
\newblock In \emph{Proceedings of EMNLP}.

\bibitem[{Vig et~al.(2020)Vig, Gehrmann, Belinkov, Qian, Nevo, Singer, and
  Shieber}]{vig2020causal}
Jesse Vig, Sebastian Gehrmann, Yonatan Belinkov, Sharon Qian, Daniel Nevo,
  Yaron Singer, and Stuart Shieber. 2020.
\newblock Investigating gender bias in language models using causal mediation
  analysis.
\newblock In \emph{Advances in Neural Information Processing Systems
  (NeurIPS)}.

\bibitem[{Yang et~al.(2024)Yang, Wang, Lu, Liu, Le, Zhou, and
  Chen}]{yang2024opro}
Chengrun Yang, Xuezhi Wang, Yifeng Lu, Hanxiao Liu, Quoc~V. Le, Denny Zhou, and
  Xinyun Chen. 2024.
\newblock Large language models as optimizers.
\newblock In \emph{International Conference on Learning Representations
  (ICLR)}.

\bibitem[{Yao et~al.(2024)Yao, Shinn, Razavi, and Narasimhan}]{yao2024tau}
Shunyu Yao, Noah Shinn, Pedram Razavi, and Karthik Narasimhan. 2024.
\newblock {$\tau$-bench}: A benchmark for tool-agent-user interaction in
  real-world domains.
\newblock In \emph{Advances in Neural Information Processing Systems
  (NeurIPS)}.

\bibitem[{Yao et~al.(2023)Yao, Zhao, Yu, Du, Shafran, Narasimhan, and
  Cao}]{yao2023react}
Shunyu Yao, Jeffrey Zhao, Dian Yu, Nan Du, Izhak Shafran, Karthik Narasimhan,
  and Yuan Cao. 2023.
\newblock {ReAct}: Synergizing reasoning and acting in language models.
\newblock In \emph{International Conference on Learning Representations
  (ICLR)}.

\bibitem[{Yuksekgonul et~al.(2024)Yuksekgonul, Bianchi, Boen, Liu, Huang,
  Guestrin, and Zou}]{yuksekgonul2024textgrad}
Mert Yuksekgonul, Federico Bianchi, Joseph Boen, Sheng Liu, Zhi Huang, Carlos
  Guestrin, and James Zou. 2024.
\newblock {TextGrad}: Automatic ``differentiation'' via text.
\newblock \emph{arXiv preprint arXiv:2406.07496}.

\bibitem[{Zhou et~al.(2023)Zhou, Muresanu, Han, Paster, Pitis, Chan, and
  Ba}]{zhou2023ape}
Yongchao Zhou, Andrei~Ioan Muresanu, Ziwen Han, Keiran Paster, Silviu Pitis,
  Harris Chan, and Jimmy Ba. 2023.
\newblock Large language models are human-level prompt engineers.
\newblock In \emph{International Conference on Learning Representations
  (ICLR)}.

\end{thebibliography}

\appendix

\section{Preliminaries}
\label{app:prelim}

\paragraph{Multi-module pipeline.} A multi-module agent is a
pipeline $M_1 \to \cdots \to M_k$ where each $M_i$ is a separate
LLM call. We map $\tau$-bench retail tasks to
$M_1$ (query rewrite), $M_2$ (planner), $M_3$ (router),
$M_4$ (response generator), with a non-LLM tool call between
$M_3$ and $M_4$.

\paragraph{Causal interventions.} $\mathrm{do}(M_i{=}v)$
\citep{pearl2009causality} sets $M_i$'s output to $v$ and
re-executes downstream modules. Oracle output $S^*_i$ (Claude
Sonnet~4.6, cached offline on $\mathcal{D}_{\mathrm{diag}}$)
enters in two forms: (i)~single-module replacement for $\Delta F_i$;
(ii)~Pearl mediation~\citep{pearl2014interpretation} for
$\mathrm{NIE}_i$ (App.~\ref{app:nie}).

\paragraph{Notation.}
\label{app:notation}
Table~\ref{tab:notation} collects the symbols used throughout the
paper.

\begin{table*}[h]
\small\centering
\setlength{\tabcolsep}{8pt}
\renewcommand{\arraystretch}{1.15}
\begin{tabular}{ll}
\toprule
\textbf{Symbol} & \textbf{Meaning} \\
\midrule
$\mathrm{sev}_i$               & severity of module $i$ on a task, scored by the LLM judge, $\in [0, 0.99)$ \\
$F$                            & failure index of an episode, $-\sum_i \log(1{-}\mathrm{sev}_i)$ (lower is better) \\
$\Delta F_i$                   & causal contribution of $M_i$, $F-F\bigl(\mathrm{do}(M_i{=}S^*_i)\bigr)$ \\
$\mathrm{TE}^{(\mathrm{med})}_i$ & mediation total effect, $F\bigl(\mathrm{do}(M_{<i}{=}S^*_{<i})\bigr) - F(E)$ (upstream oracle $+$ $M_i$ refreshed; ${\neq}{-}\Delta F_i$) \\
$\mathrm{NDE}_i$               & natural direct effect, $F(\text{B}){-}F(E)$
                                 (see App.~\ref{app:nie}) \\
$\mathrm{NIE}_i$               & natural indirect effect (Pearl $X{=}x^*$ arm,
                                 $\mathrm{NIE}_r$), $F(\text{A}){-}F(\text{B})$
                                 (see App.~\ref{app:nie}) \\
$i^\star_{\mathrm{pop}}$       & population diagnostic target, $\arg\max_i \overline{\Delta F_i}$ \\
$\mathrm{fate}(i,c)$           & per-task role of $M_i$ on case $c$: amplifier / propagator / compensator, classified by $\mathrm{NIE}_i(c)$ \\
$\mathcal{D}_{\mathrm{diag}}$  & diagnostic split ($\tau$-bench retail train, $n_d{=}500$) \\
$\mathcal{D}_{\mathrm{presc}}$ & prescription split ($\tau$-bench retail test, $n_p{=}113$, $111$ valid) \\
CCP                            & Counterfactual Correction Patching (\S\ref{ssec:ccp}) \\
\bottomrule
\end{tabular}
\caption{Symbol legend. Three usage classes: (i)~\emph{failure} is
exactly $F$; $\Delta F_i$ and $\mathrm{NIE}_i$ are causal effects on
$F$, not failures. (ii)~\emph{Diagnosis} has two levels: population
$i^\star_{\mathrm{pop}}$ and per-task $\mathrm{fate}(i,c)$.
(iii)~\emph{Prescription} refers to modifying the agent
(rewrite, compute upgrade, CCP, oracle injection);
\emph{intervention} refers to causal $\mathrm{do}(\cdot)$
operations; \emph{patch} refers specifically to the CCP
correction mechanism.}
\label{tab:notation}
\end{table*}

\section{NIE: Detailed Definition, Worked Example, and Sign Analysis}
\label{app:nie}

\paragraph{Detailed definition.} Let $M_i^{(0)}$ denote $M_i$'s output
in the baseline run (under noisy upstream), and let $M_i(S^*_{<i})$
denote $M_i$'s output when re-executed on oracle upstream values for
$M_{<i}$. $\mathrm{NIE}_i$ compares two intervened worlds:
\begin{itemize}\itemsep=0pt
\item \textbf{World A (refresh):} replace upstream with oracle
  ($\mathrm{do}(M_{<i}{=}S^*_{<i})$), re-execute $M_i$ on the new
  upstream, then re-execute $M_{>i}$. Yields $F(\text{A})$.
\item \textbf{World B (frozen):} same upstream replacement, but hold
  $M_i$ at its original baseline output ($M_i^{(0)}$); only $M_{>i}$
  is re-executed. Yields $F(\text{B})$.
\end{itemize}
$\mathrm{NIE}_i = F(\text{A}) - F(\text{B})$. The two worlds share
the oracle upstream and the same downstream re-execution; the only
difference is whether $M_i$ adapted to the cleaner upstream or kept
its baseline-derived output. The mediation
identity~\citep{pearl2014interpretation} gives
$\mathrm{TE}^{(\mathrm{med})}_i = \mathrm{NDE}_i + \mathrm{NIE}_i$, where
$\mathrm{TE}^{(\mathrm{med})}_i = F(\text{A}) - F(E)$ and
$\mathrm{NDE}_i = F(\text{B}) - F(E)$.

\paragraph{Sign analysis (per task).}
\begin{itemize}\itemsep=0pt
\item $\mathrm{NIE}_i > +\tau$ -- \textsc{amplifier}: $M_i(S^*_{<i})$
  is structurally cleaner than $M_i^{(0)}$ but lies outside the
  distribution that downstream $M_{>i}$ was tuned for, so $F$ rises.
  This is the cascade signature.
\item $\mathrm{NIE}_i < -\tau$ -- \textsc{compensator}: $M_i(S^*_{<i})$
  is in-family for downstream and improves the final outcome; $M_i$
  tracks upstream quality.
\item $|\mathrm{NIE}_i| \le \tau$ -- \textsc{propagator}: $M_i$'s
  output is effectively invariant to upstream cleanup at this task.
\end{itemize}

\paragraph{Identification note.}
Our $\mathrm{NIE}_i = F(\text{A}) - F(\text{B})$ corresponds to the
$X{=}x^*$ arm of Pearl's mediation decomposition ($\mathrm{NIE}_r$),
which together with $\mathrm{NDE}_i = F(\text{B}) - F(E)$ satisfies
$\mathrm{TE}^{(\mathrm{med})}_i = \mathrm{NDE}_i + \mathrm{NIE}_i$
by telescoping. Equivalence with the standard
$\mathrm{NIE} = Y(x, M(x^*)) - Y(x, M(x))$ requires no upstream
${\times}$ $M_i$ interaction on $F$, i.e., $F$ factorizes as
$f(M_{<i}) + g(M_i, M_{>i})$ up to noise. The chain structure of our
pipeline naturally motivates this, with $\mathrm{NDE}_i {\neq} 0$ in
Table~\ref{tab:e1e2} reflecting residual cross-prompt context sharing
(e.g., $M_4$ accessing the user query alongside $M_3$'s output) and
stale-output mismatch in World~B. The fate signs of
\S\ref{ssec:nie-method} are defined directly on $\mathrm{NIE}_r$ and
remain valid regardless.

\section{CCP Correction Pool Construction}
\label{app:ccp-pool}

The correction pool used by every CCP configuration is built
once, offline on $\mathcal{D}_{\mathrm{diag}}$ ($n_d{=}500$),
before any test-time evaluation on $\mathcal{D}_{\mathrm{presc}}$.
Source: the intervention sweep we already run for $\Delta F$
identification (\S\ref{ssec:paradox}), which produces, for every diagnosis
task $c \in \mathcal{D}_{\mathrm{diag}}$ and every module $M_i$, the
triple $(\text{module input}, M_i^{(0)}(c),\ S^*_i(c))$ where
$M_i^{(0)}(c)$ is the agent's baseline output and $S^*_i(c)$ is the
cached Sonnet 4.6 oracle output, together with the LLM-judge severity
$\mathrm{sev}_i^{(0)}(c) \in [0, 0.99]$ scoring how badly
$M_i^{(0)}(c)$ misses $S^*_i(c)$.

\textbf{Selection rule.} For each module $M_i$ we restrict candidates
to tasks with $\mathrm{sev}_i^{(0)}(c) \ge \tau_{\mathrm{demo}}$ (we
use $\tau_{\mathrm{demo}}{=}0.30$, the judge's
``partial coverage / missing one minor element'' rubric anchor),
then sort by $\mathrm{sev}_i^{(0)}(c)$ descending and take the top
$k{=}5$ (ties broken by task id). The earlier raw-string criterion
$M_i^{(0)} \neq S^*_i$ admitted paraphrase-only differences (e.g.,
$M_1$ third-person rewrite, $M_4$ verbose reply) which carry no learning
signal; the severity-based rule selects only triples where the
judge has confirmed a semantic gap.

\textbf{Patch template.} Each correction renders into the module's
prompt as a single block of the form
\texttt{Input: <text>~|~Wrong: <text>~|~Correct: <text>}; instructions
and the output schema of $M_i$ are not modified. The only difference
between population and per-task CCP configurations is which
module's pool is selected at test time. No task in
$\mathcal{D}_{\mathrm{presc}}$ contributes to any pool, so test-time
patching never observes the task it is being applied to.

\section{CCP Procedures}
\label{app:a6-proc}

\paragraph{Routing targets.} Pop CCP and its naive
symptom-axis counterpart route to a \emph{single} module index
fixed across all test tasks. The CICA population $\Delta F$
target is
\begin{equation}
i^\star_{\mathrm{pop}}
\;=\;
\arg\max_{i\,\in\,\{M_1,\ldots,M_4\}}\,
\frac{1}{N}\sum_{c\,\in\,\mathcal{D}}
\Delta F_i(c),
\label{eq:ipop}
\end{equation}
the module with the largest mean
$\Delta F_i$ (Eq.~\ref{eq:deltaF}) on the diagnosis split
$\mathcal{D}_{\mathrm{diag}}$ ($n_d{=}500$);
$i^\star_{\mathrm{pop}}$ is computed once at diagnosis and reused
on every $c \in \mathcal{D}_{\mathrm{presc}}$. The naive
symptom counterpart is
\begin{equation}
i^{\mathrm{pop\text{-}sev}}
\;=\;
\arg\max_{i}\,
\frac{1}{N}\sum_{c}\mathrm{sev}_i^{(0)}(c),
\label{eq:i-popsev}
\end{equation}
where $\mathrm{sev}_i^{(0)}(c)$ is the LLM-judge severity of
$M_i$ on the baseline run of task $c$. The naive per-task
severity target $i^{\mathrm{task\text{-}sev}}(c) =
\arg\max_i \mathrm{sev}_i^{(0)}(c)$ and the $z$-score adaptive
target (\S\ref{ssec:ccp}) are recomputed per task and used by the
adaptive procedures.

\begin{algorithm}[t]
\caption{Pop CCP at the population $\Delta F$ target
$i^\star_{\mathrm{pop}}$.}
\label{alg:a6}
\begin{algorithmic}[1]
\Require correction pools $\{P_i\}$ for every module
  (App.~\ref{app:ccp-pool}); population $\Delta F$ target
  $i^\star_{\mathrm{pop}} = \arg\max_i \overline{\Delta F_i}$
  (Eq.~\ref{eq:ipop}); task with cached oracle $S^*$
\State Patch $M_{i^\star_{\mathrm{pop}}}$'s prompt with $k$ corrections
  drawn from $P_{i^\star_{\mathrm{pop}}}$; other modules unchanged
\State Run the pipeline $M_1 \to M_2 \to M_3 \to M_4$
\State $F \gets$ LLM judge against $S^*$
\State \Return $F$
\end{algorithmic}
\end{algorithm}

This procedure makes no test-time decisions and never calls a
routing oracle; the same $k$-correction patch is applied to
$M_{i^\star_{\mathrm{pop}}}$ on every task. The Naive Pop
Severity configuration follows the identical procedure but with
$i^{\mathrm{pop\text{-}sev}}$ (Eq.~\ref{eq:i-popsev}) replacing
$i^\star_{\mathrm{pop}}$.

The four matched-treatment configurations run exactly the
procedure above with the target module fixed to $M_1$, $M_2$,
$M_3{=}i^\star_{\mathrm{pop}}$, and $M_4$ respectively, with
$k$, pool-construction rule, correction order, and judge held
identical across the four. This is the design that makes the
four-point Pop CCP measurement (App.~\ref{app:cross-agent}) a
valid isolation of the target-module choice.

\paragraph{Adaptive $z$-score routing variant.}
\label{app:a8-proc}
The Adaptive CCP configuration routes the patch per task by an
at-test-time estimate of $\mathrm{NIE}_i$, using a cheap
secondary LLM (Gemma-4-26B-A4B-it) to generate a fresh
per-task routing oracle. The correction pools $\{P_i\}$
themselves are built once on $\mathcal{D}_{\mathrm{diag}}$
(App.~\ref{app:ccp-pool}); only the routing decision is
task-adaptive.

\begin{algorithm}[t]
\caption{Adaptive CCP via per-task $\mathrm{NIE}_i$ routing.}
\label{alg:a8}
\begin{algorithmic}[1]
\Require correction pools $\{P_i\}$ from $\mathcal{D}_{\mathrm{diag}}$;
  cheap routing-oracle LLM (Gemma); LLM judge; threshold $\tau$;
  test task $c \in \mathcal{D}_{\mathrm{presc}}$ with cached scoring
  oracle $S^*(c)$
\State Run the unmodified pipeline on $c$ $\to$ baseline state $\sigma_0$
\State $\hat S^* \gets \text{Gemma}(user\_query, \sigma_0)$
  \hfill // routing oracle, deployment-realistic
\For{each module $M_i$ with $i \ge 2$}
  \State $\sigma_A \gets$ refresh $M_i$ on
    $\mathrm{do}(M_{<i}{=}\hat S^*_{<i})$, re-run downstream
  \State $F_A \gets$ LLM judge on $\sigma_A$ vs.\ $\hat S^*$
  \State $\sigma_B \gets$ freeze $M_i$ at $\sigma_0$'s value under
    $\mathrm{do}(M_{<i}{=}\hat S^*_{<i})$, re-run downstream
  \State $F_B \gets$ LLM judge on $\sigma_B$ vs.\ $\hat S^*$
  \State $\mathrm{NIE}_i \gets F_A - F_B$
\EndFor
\State $\hat\imath^* \gets \arg\max_i \mathrm{NIE}_i$
\If{$\mathrm{NIE}_{\hat\imath^*} > \tau$}
  \State Patch $M_{\hat\imath^*}$'s prompt with $k$ corrections from
    $P_{\hat\imath^*}$; re-run the pipeline $\to \sigma_{\text{out}}$
\Else
  \State $\sigma_{\text{out}} \gets \sigma_0$ \hfill // no patch
\EndIf
\State $F \gets$ LLM judge on $\sigma_{\text{out}}$ vs.\ $S^*(c)$
  \hfill // scored against the strong cached oracle
\State \Return $F$, $\hat\imath^*$
\end{algorithmic}
\end{algorithm}

\textbf{Per-task cost} (gpt-4o-mini agent, Gemini~2.5~Flash judge):
The adaptive configuration adds 1 Gemma call (routing oracle),
$2{\times}3 = 6$ intervention re-runs, and 7 judge scorings on
top of the baseline run.

\textbf{Why two oracles, and why this is not leakage.} The routing
oracle $\hat S^*$ is generated at test time from a cheap LLM and is
used only to decide \emph{which module} to patch; the patch itself
draws from $P_i$ built on $\mathcal{D}_{\mathrm{diag}}$. The cached
Sonnet 4.6 oracle $S^*(c)$ is used only at the final scoring step
(line~$F \gets \dots$) and is never seen by the routing decision or
by the agent. Hence the routing rule is deployment-implementable
(no privileged information about $c$ beyond what Gemma can infer
from the live query) while scoring fidelity is preserved.

The $z$-score variant replaces lines~9--10 of
Algorithm~\ref{alg:a8} with the $z$-score selection rule. Let $(\mu_i, \sigma_i)$ be the per-module
$\mathrm{NIE}$ mean and standard deviation computed once on
$\mathcal{D}_{\mathrm{diag}}$ (cached alongside $\{P_i\}$). At test
time:
\[
z_i(c) \,=\, \frac{\mathrm{NIE}_i(c) - \mu_i}{\sigma_i},\qquad
\hat\imath^* \,=\, \arg\max_i z_i(c),
\]
with the patch fired iff $z_{\hat\imath^*}(c) > z_{\mathrm{thr}}$
(default $z_{\mathrm{thr}}{=}1.0$). The Pearl-mediation sign of
$\mathrm{NIE}_i$ is preserved as an absolute quantity for fate
labelling (\S\ref{ssec:nie-method}); the $z$-score is used only
for routing, where the cross-module comparison needs to account
for $\sigma_{\mathrm{NIE}_i}$'s $3{\times}$ variation across
modules (Table~\ref{tab:e1e2}). All other steps of Algorithm~\ref{alg:a8} are
unchanged.

\section{Full Configuration Specification}
\label{app:configurations}

Table~\ref{tab:configurations-full} lists the 11 prescription
configurations grouped by their role in the paper: the seven
\emph{Main} configurations carry the lead claim, and the four
\emph{Ablation} configurations isolate the treatment-family axis
at $i^\star_{\mathrm{pop}}$.

\begin{table*}[h]
\small\centering
\setlength{\tabcolsep}{8pt}
\renewcommand{\arraystretch}{1.18}
\begin{tabular}{@{}l l l@{}}
\toprule
\textbf{Group} & \textbf{Target} & \textbf{Treatment} \\
\midrule
\multicolumn{3}{l}{\emph{Main: four-point Pop CCP + adaptive variant + UB}} \\
Baseline   & --                                                       & --              \\
Pop CCP    & $M_1$ (QR)                                               & Pop CCP         \\
           & $M_2$ (Planner)                                          & Pop CCP         \\
           & $M_3$ ($i^\star_{\mathrm{pop}}$)                         & Pop CCP         \\
           & $M_4$ (RG)                                               & Pop CCP         \\
Adaptive   & per-task $\arg\max_i z_i(c)$                             & CCP             \\
Upper bound & $M_3$ (oracle inject)                                   & oracle $S^*_i$  \\
\midrule
\multicolumn{3}{l}{\emph{Ablation: alternate treatments at $M_3$}} \\
           & $M_3$                                                    & Rewrite         \\
           & $M_3$                                                    & Compute upgrade \\
           & per-task $\arg\max_i \mathrm{NIE}_i$ (absolute)           & CCP             \\
\bottomrule
\end{tabular}
\caption{Full configuration specification. The \emph{Main} block
carries the Diagnostic Paradox finding; the \emph{Ablation} block
shows that the $M_3$ Hazard is \emph{CCP-specific}: non-CCP
treatments at $M_3$ are all neutral, confirming the Hazard
requires precise output-distribution shift; absolute-NIE routing
is dominated by its $z$-score variant.}
\label{tab:configurations-full}
\end{table*}

Table~\ref{tab:pvalues} reports raw and Holm--Bonferroni-corrected
$p$-values for the four critical comparisons that establish the
Diagnostic Paradox.

\begin{table}[h]
\small\centering
\setlength{\tabcolsep}{6pt}
\renewcommand{\arraystretch}{1.15}
\begin{tabular}{lccc}
\toprule
\textbf{Comparison} & \textbf{Raw $p$} & \textbf{Holm $p$} & \textbf{$d_z$} \\
\midrule
Baseline vs.~$M_1$            & $0.043$ & $0.086$ & $-0.18$ \\
Baseline vs.~$M_3$            & $0.012$ & $0.036^{\ast}$ & $+0.22$ \\
$M_1$ vs.~$M_3$ (primary)     & $0.002$ & $0.008^{\ast\ast}$ & $+0.37$ \\
$M_3$ Pop CCP vs.~$z$-score   & $0.003$ & $0.009^{\ast\ast}$ & $-0.27$ \\
\bottomrule
\end{tabular}
\caption{Holm--Bonferroni corrected $p$-values (two-sided paired
Wilcoxon, $n_p{=}111$). $^{\ast}p{<}0.05$, $^{\ast\ast}p{<}0.01$.
The baseline-vs.-$M_1$ leverage comparison does not survive Holm
correction; the paradox is established by the $M_1$ vs.~$M_3$
contrast.}
\label{tab:pvalues}
\end{table}

\section{Models and Roles}
\label{app:models}

Table~\ref{tab:models} lists the models used in all experiments, their families, and their roles in the pipeline.

\begin{table*}[h]
\small\centering
\setlength{\tabcolsep}{6pt}
\renewcommand{\arraystretch}{1.15}
\begin{tabular}{|l|l|l|p{0.42\linewidth}|}
\hline
\textbf{Role} & \textbf{Model} & \textbf{Family} & \textbf{Used by / when} \\
\hline
Agent & gpt-4o-mini ($T{=}0$)   & OpenAI    & Main experiment (full 7-configuration sweep, $n_d{=}500$, $n_p{=}111$). \\
Agent & Llama~4 Scout ($T{=}0$) & Meta      & Cross-agent replication; full 7-configuration sweep, same splits. \\
Agent & Qwen3-32b ($T{=}0$)     & Alibaba   & Prescription pilot (baseline + Pop CCP @ $M_1$/$M_3$, $n_p{=}49$) and diagnostic sweep ($n{=}50$ stratified). \\
Diagnosis oracle  & Claude Sonnet 4.6       & Anthropic & Diagnosis only, offline. Produces per-stage gold $S^*_i$ once per task; cached and reused for every prescription configuration. \\
LLM judge         & Gemini 2.5 Flash        & Google    & Diagnosis and prescription, every (case, module). Scores $\mathrm{sev}_i \in [0,0.99]$ by comparing $S_i$ to $S^*_i$. \\
Routing oracle    & Gemma-4-26B-A4B-it      & Google    & Adaptive variants only, test time. Generates per-task $\hat S^*$ for computing $\mathrm{NIE}_i$ used to route the patch. \\
Rewrite optimizer & GPT-5                   & OpenAI    & Pop Rewrite configuration only. Sees $\le 8$ (input, output, $S^*$) examples for the target module; emits a new system prompt held fixed at test time. \\
Compute upgrade   & Gemini 2.5 Pro          & Google    & Pop Compute configuration only. Replaces the base model of $i^\star_{\mathrm{pop}}$; other three modules remain gpt-4o-mini. \\
\hline
\end{tabular}
\caption{Models and their roles. Three distinct base agents span
OpenAI, Meta, and Alibaba families. The cross-family
oracle/judge triple \{Anthropic, Google\} prevents
self-correlation with any of the agents.
The Pop Rewrite configuration (GPT-5) produces $\Delta{=}{-}0.014$
and does not affect the main Diagnostic Paradox finding.}
\label{tab:models}
\end{table*}

\section{Airline Domain: Diagnostic Replication}
\label{app:airline-e2}

To probe whether $M_3$ (Router) is the diagnostic bottleneck
beyond $\tau$-bench retail, we ran the same E2 causal sweep
(gpt-4o-mini agent, Claude Sonnet~4.6 oracle, Gemini~2.5~Flash
judge) on the $\tau$-bench airline test split ($n{=}37$ valid
tasks). Table~\ref{tab:airline-e2} reports per-module
$\overline{\Delta F_i}$ and fate distributions.

\begin{table}[h]
\small\centering
\setlength{\tabcolsep}{6pt}
\renewcommand{\arraystretch}{1.15}
\begin{tabular}{lcccc}
\toprule
\textbf{Module} & \multicolumn{2}{c}{$\overline{\Delta F_i}$} & \multicolumn{2}{c}{\textbf{Comp\,/\,Prop\,/\,Amp}} \\
\cmidrule(lr){2-3}\cmidrule(lr){4-5}
 & \textbf{Retail} & \textbf{Airline} & \textbf{Retail} & \textbf{Airline} \\
\midrule
$M_1$ (QR)       & 0.357 & 0.207 & n/a               & n/a \\
$M_2$ (Planner)  & 0.817 & 1.453 & 454/\phantom{0}6/40 & 37/0/0 \\
$M_3$ (Router)$^\dagger$  & \textbf{1.018} & \textbf{1.690} & 491/\phantom{0}7/\phantom{0}2  & 21/12/4 \\
$M_4$ (RG)       & 0.589 & 0.664 & 492/\phantom{0}8/\phantom{0}0  & 37/0/0 \\
\bottomrule
\end{tabular}
\caption{E2 diagnostic replication on $\tau$-bench airline
($n{=}37$). $^\dagger$ $M_3$ is the largest-$\overline{\Delta F_i}$
module in both domains. Fate counts: Compensator/Propagator/Amplifier
under $|\mathrm{NIE}_i|{>}0.05$. Retail column reproduces
Table~\ref{tab:e1e2} for reference.}
\label{tab:airline-e2}
\end{table}

$M_3$ has the highest $\overline{\Delta F_i}$ in both domains
($0.357{\to}1.018$ retail; $0.207{\to}1.690$ airline), and
compensator-majority fate distributions hold for all downstream
modules. The airline amplifier count at $M_3$ (4/37, 10.8\%) is
higher than retail (2/500, 0.4\%), consistent with the harder
airline benchmark ($\bar F{=}3.63$ vs.~$2.45$ retail).

\section{Cross-Agent Replication}
\label{app:cross-agent}

\paragraph{Setup.}
All settings identical to \S\ref{ssec:setup} except the base
agent is replaced by Llama~4 Scout ($T{=}0$). Oracle (Claude
Sonnet~4.6) and judge (Gemini~2.5~Flash) are unchanged.
$n_p{=}111$ prescription tasks (same split as gpt-4o-mini).

\paragraph{Diagnostic replication.}
$M_3$ is the population diagnostic bottleneck
($\overline{\Delta F_3}{=}1.564$, highest across modules),
replicating the gpt-4o-mini diagnosis ($1.018$). Full prescription
results across all three agents are in Table~\ref{tab:main-result}.
Notably, $M_3$ compensator rate is $98.2\%$ (gpt-4o-mini), $96.0\%$
(Qwen3-32b), and $0\%$ (Llama~4 Scout), consistent with the
Hazard trend: absolute Hazard appears on the two high-rate agents
and is absent on Llama~4 Scout.

\begin{table*}[h]
\centering\small
\setlength{\tabcolsep}{7pt}
\renewcommand{\arraystretch}{1.18}
\begin{tabular}{@{}l r r r@{}}
\toprule
\textbf{Configuration}
  & \textbf{gpt-4o-mini} ($n{=}111$)
  & \textbf{Llama~4 Scout} ($n{=}111$)
  & \textbf{Qwen3-32b} ($n{=}49$) \\
\midrule
\multicolumn{4}{l}{\emph{All three agents}} \\
Baseline $F$ (tool-match)
  & 3.687 (37.8\%)
  & 3.601 (42.3\%)
  & 3.533 (44.9\%) \\
Pop CCP @ $M_1$ ($\Delta$)
  & $\boldsymbol{-0.191}$ (44.1\%)
  & $\boldsymbol{-0.402}$ (53.2\%)
  & $\boldsymbol{-0.659}$ (44.9\%) \\
Pop CCP @ $M_3$ ($\Delta$)
  & $+0.243^{\ast}$ (33.3\%)
  & $-0.013$
  & $+0.683^{\ast\ast\ast}$ (24.5\%) \\
$M_1$-vs.-$M_3$ contrast
  & $+0.435^{\ast\ast}$
  & $+0.389^{\dagger}$
  & $+1.343^{\ast\ast\ast}$ \\
$M_3$ compensator rate
  & 98.2\%  & 0\%  & 96.0\% \\
\midrule
\multicolumn{4}{l}{\emph{gpt-4o-mini and Llama~4 Scout only (full configuration sweep)}} \\
Pop CCP @ $M_2$ ($\Delta$)  & $-0.043$ & $-0.053$ & \\
Pop CCP @ $M_4$ ($\Delta$)  & $-0.024$ & $-0.094$ & \\
Adaptive $z$-score ($\Delta$) & $-0.029$ & $-0.137^{\ast\ast}$ & \\
Oracle upper bound ($\Delta$) & $-2.007^{\ast\ast\ast}$ & $-1.795^{\ast\ast\ast}$ & \\
\bottomrule
\end{tabular}
\caption{Prescription $\Delta$ (change from baseline; negative = improvement) and
tool-match rate for all three agents. $M_3$-worst, $M_1$-best ordering holds
in every agent. Absolute Hazard ($\Delta(M_3){>}0$) appears on the two
high-compensator agents (gpt-4o-mini, Qwen3-32b) but not on Llama~4 Scout,
consistent with the Contract trend.
$M_3$ tool-match shows the clearest behavioral signal of Hazard (drops from
37.8\% to 33.3\% on gpt-4o-mini, 44.9\% to 24.5\% on Qwen3-32b).
Paired Wilcoxon vs.\ baseline.
$^{\dagger}$ directional only (raw $p{=}0.059$, Holm $p{=}0.177$);
$^{\ast}$ Holm $p{<}0.05$;
$^{\ast\ast}$ Holm $p{<}0.01$;
$^{\ast\ast\ast}$ raw $p{<}0.001$.
Qwen3-32b uses gpt-4o-mini correction triples; `—' = configuration not tested.}
\label{tab:main-result}
\end{table*}

\section{CCP Correction-Pool Examples}
\label{app:ccp-examples}

Each correction in the per-module pool is a triple
$(x_i,\ S_i^{\mathrm{actual}},\ S^*_i)$ extracted at zero
additional inference cost from the diagnosis sweep on
$\mathcal{D}_{\mathrm{diag}}$: $x_i$ is what $M_i$ saw,
$S_i^{\mathrm{actual}}$ is what $M_i$ produced in the baseline
run, and $S^*_i$ is the cached oracle output. Selection rule:
$\mathrm{sev}_i(c) \ge \tau_{\mathrm{demo}}{=}0.30$, top-$k{=}5$
by descending severity. The actual correction that lands in
$M_i$'s prompt renders as
{\small\ttfamily \#\#\# Example $j$ / Input: $\{x_i\}$ /
\xmark\ Wrong: $\{S_i^{\mathrm{actual}}\}$ /
\cmark\ Correct: $\{S^*_i\}$}.
Figure~\ref{fig:ccp-pool-examples} shows one real top-severity
correction per module from the $\mathcal{D}_{\mathrm{diag}}$
($n_d{=}500$) pool used in all CCP configurations in this paper;
user queries are shortened for legibility, the rest are verbatim.

\begin{figure*}[!htbp]
\centering
\setlength{\fboxsep}{6pt}
\renewcommand{\arraystretch}{1.15}
\small
\begin{tabular}{@{}p{0.46\textwidth}@{\hspace{0.01\textwidth}}|@{\hspace{0.01\textwidth}}p{0.46\textwidth}@{}}
\toprule
\textbf{$M_1$ (Query Rewrite) ---
\texttt{tau\_retail\_008}, $\mathrm{sev}_1{=}0.70$} &
\textbf{$M_2$ (Planner) ---
\texttt{tau\_retail\_043}, $\mathrm{sev}_2{=}0.70$} \\
\midrule
\textbf{Input (\texttt{user\_query}):}
``My name is Lucas Muller and my email is
lucas.muller7899@example.com. I'm patient, cautious. For
\#W3206099, modify the Gaming Mouse \ldots''
\par\smallskip
\xmark\ \textbf{Wrong (\texttt{rewritten\_query}):}
``\ldots\ For order \#W3206099, modify the Gaming Mouse with
color black, sensor \ldots''
\par\smallskip
\cmark\ \textbf{Oracle (\texttt{rewritten\_query}):}
``\ldots\ For order W3206099, I want to modify the Gaming Mouse
with color black \ldots''
\par\smallskip
\textit{Failure type:} preserved the \texttt{\#} sigil and copied
the listed adjectives verbatim instead of normalizing the
request into an action-oriented rewrite.
&
\textbf{Input (rewritten query from $M_1$):}
``My name is Lei Ahmed \ldots Cancel order \#W9132\ldots''
\par\smallskip
\xmark\ \textbf{Wrong (\texttt{intent}):}
\texttt{modify\_request}
\par\smallskip
\cmark\ \textbf{Oracle (\texttt{intent}):}
\texttt{cancel\_request}
\par\smallskip
\textit{Failure type:} intent classification confused the
explicit ``cancel'' verb with a modification request, plausibly
because the user also mentions other modifications later in the
message.
\\
\midrule
\textbf{$M_3$ (Router) ---
\texttt{tau\_retail\_035}, $\mathrm{sev}_3{=}0.95$} &
\textbf{$M_4$ (Response Generator) ---
\texttt{tau\_retail\_059}, $\mathrm{sev}_4{=}0.70$} \\
\midrule
\textbf{Input (planner state):}
intent~$=\!$\,\texttt{modify\_request},
slots~$=\!$\,\{order\_id, address fields, item modifications\};
user explicitly asks to ``change the shipping address''.
\par\smallskip
\xmark\ \textbf{Wrong (\texttt{tool\_name}):}
\texttt{modify\_pending\_order\_items}
\par\smallskip
\cmark\ \textbf{Oracle (\texttt{tool\_name}):}
\texttt{modify\_pending\_order\_address}
\par\smallskip
\textit{Failure type:} router collapsed two distinct
modification verbs (address vs.\ items) into the more frequent
action. Highest-severity tier in the pool ($0.95$); $M_3$'s pool
is dominated by this kind of tool mis-selection.
&
\textbf{Input (router state + tool result):}
exchange of Makeup Kit on order W6131421 with payment method
substitution.
\par\smallskip
\xmark\ \textbf{Wrong (\texttt{final\_response}):}
``Hello Anya Patel, thank you for your request to exchange your
Makeup Kit. I will assist you in exchanging the kit for the
desired skin tone \ldots''
\par\smallskip
\cmark\ \textbf{Oracle (\texttt{final\_response}):}
``Hi Anya, I'm processing the exchange for the Makeup Kit in
order W6131421 (switching to skin tone `dark', Brand~B) via your
credit card ending\ldots''
\par\smallskip
\textit{Failure type:} generically polite reply but omits the
concrete order ID, the new skin tone, and the payment-method
substitution that the tool actually performed.
\\
\bottomrule
\end{tabular}
\caption{One top-severity CCP correction per module, drawn from
the $\mathcal{D}_{\mathrm{diag}}$ ($n_d{=}500$) pool. The four
examples illustrate the failure pattern each module's pool
encodes: surface formatting ($M_1$), intent classification
($M_2$), tool selection ($M_3$, the highest-severity tier), and
concrete-field omission ($M_4$). When CCP applies at $M_3$ (the
diagnosed target), all five corrections are high-severity
tool mis-selections; the patched prompt shifts $M_3$'s
tool-name distribution sharply --- exactly the
output-distribution shift the Hazard mechanism identifies in
\S\ref{ssec:mechanism}.}
\label{fig:ccp-pool-examples}
\end{figure*}

Figure~\ref{fig:causal-rejection-cases} shows three $M_2$-amplifier
cases that make the Linguistic Contract concrete: in each, the baseline
$M_2$ output is less informative than the oracle's, yet the downstream
recovers because $M_3/M_4$ have absorbed the surface cues of the noisy
upstream.

\begin{figure*}[!htbp]
\centering\small
\fbox{\begin{minipage}{0.98\textwidth}
\textbf{Case~1}~~\texttt{tau\_retail\_243}~~($\mathrm{NIE}_{M_2}{=}{+}1.099$)\\[2pt]
\emph{User query:}~``\ldots Cancel order \#W7807323 because ordered by mistake.
For \#W2618034, change address to \ldots''\\[3pt]
\begin{tabular}{@{}p{0.30\textwidth} p{0.30\textwidth} p{0.34\textwidth}@{}}
\toprule
\textbf{Baseline $M_2$} ($M_2^{(0)}$) & \textbf{Oracle $M_2$} ($S_2^*$) & \textbf{Baseline downstream} ($M_3^{(0)}, M_4^{(0)}$) \\
\midrule
\texttt{intent=modify\_request} & \texttt{intent=cancel\_request} & $M_3$: chooses \texttt{cancel\_pending\_order} \\
slots: \{name, email, request\}  & slots: $+$ explicit \texttt{order\_id} & $M_4$: confirms the cancellation \\
rationale: ``multiple requests for cancellation and modifications'' & rationale: ``cancellation of W7807323 due to ordering by mistake \ldots primary intent'' & ~ \\
\bottomrule
\end{tabular}\\[3pt]
\emph{Mechanism:} The contract holds because $M_3$ keys off the
literal user-query verb ``Cancel'' rather than $M_2$'s
mis-labelled intent. Substituting the oracle $M_2$ output would
remove this surface cue from the upstream channel that $M_3$
relies on.
\end{minipage}}\\[6pt]
\fbox{\begin{minipage}{0.98\textwidth}
\textbf{Case~2}~~\texttt{tau\_retail\_017}~~($\mathrm{NIE}_{M_2}{=}{+}1.099$)\\[2pt]
\emph{User query:}~``\ldots For \#W6026015, exchange Luggage Set
\{piece count: 2-piece, color: red, material: hardshell\} to
\{material: softshell\}\ldots''\\[3pt]
\begin{tabular}{@{}p{0.30\textwidth} p{0.30\textwidth} p{0.34\textwidth}@{}}
\toprule
\textbf{Baseline $M_2$} ($M_2^{(0)}$) & \textbf{Oracle $M_2$} ($S_2^*$) & \textbf{Baseline downstream} ($M_3^{(0)}, M_4^{(0)}$) \\
\midrule
\texttt{intent=exchange\_request} & \texttt{intent=exchange\_request} & $M_3$: \texttt{exchange\_delivered\_order\_items} with item IDs as free-text names (\texttt{"Luggage Set"}, \ldots) and \texttt{payment\_method\_id}~$=$~user email \\
rationale: \emph{empty}            & rationale: ``two item exchanges on W6026015 with specific attribute changes and a named payment method'' & $M_4$: confident anticipatory response (``I will proceed with \ldots'') \\
\bottomrule
\end{tabular}\\[3pt]
\emph{Mechanism:} The downstream prompt has absorbed
gpt-4o-mini's habit of returning item names instead of catalog
IDs and routes the response phrasing accordingly. Substituting
the oracle $M_2$ would not fix $M_3$'s ID-vs-name habit but
would change $M_2$'s phrasing, which is what downstream had been
keying off.
\end{minipage}}\\[6pt]
\fbox{\begin{minipage}{0.98\textwidth}
\textbf{Case~3}~~\texttt{tau\_retail\_036}~~($\mathrm{NIE}_{M_2}{=}{+}1.099$)\\[2pt]
\begin{tabular}{@{}p{0.30\textwidth} p{0.30\textwidth} p{0.34\textwidth}@{}}
\toprule
\textbf{Baseline $M_2$} ($M_2^{(0)}$) & \textbf{Oracle $M_2$} ($S_2^*$) & \textbf{Baseline downstream} ($M_3^{(0)}, M_4^{(0)}$) \\
\midrule
\texttt{intent=modify\_request} & \texttt{intent=modify\_request} & $M_3$: \texttt{modify\_pending\_order\_items} with descriptive item names \\
rationale: \emph{empty}         & rationale: rich                 & $M_4$: ``I will proceed with updating the Espresso Machine \ldots'' \\
\bottomrule
\end{tabular}\\[3pt]
\emph{Mechanism:} Same compensation pattern as Case~2 ---
downstream relies on the surface phrasing of the noisy
baseline $M_2$.
\end{minipage}}
\caption{Three $M_2$-amplifier cases (three highest-$\mathrm{NIE}_{M_2}$ tasks in $\mathcal{D}_{\mathrm{diag}}$)
drawn from $\mathcal{D}_{\mathrm{diag}}$. Each row juxtaposes the Planner's noisy baseline
output ($M_2^{(0)}$), its oracle-quality counterpart ($S_2^*$), and the actual
downstream response. The downstream produces a passable response despite
the upstream's flaws, consistent with the Linguistic Contract
(\S\ref{ssec:contract}). Substituting $S_2^*$ for $M_2^{(0)}$ would
strip the absorbed surface cues and trigger the Hazard observed at
Pop CCP $M_3$ (App.~\ref{app:cross-agent}).}
\label{fig:causal-rejection-cases}
\end{figure*}

\section{$\tau$-Sensitivity of Fate Counts}
\label{app:tau-sensitivity}

The fate threshold $\tau{=}0.05$ used throughout
\S\ref{ssec:nie-method} and \S\ref{ssec:paradox} is a small-effect
convention near the LLM judge's smallest non-zero severity tier
($-\log(1{-}0.05){\approx}0.051$). Table~\ref{tab:tau-sensitivity}
reports the per-module amp/prop/comp counts under
$\tau \in \{0.01, 0.05, 0.10, 0.20\}$.

\begin{table}[h]
\centering\small
\setlength{\tabcolsep}{4pt}
\renewcommand{\arraystretch}{1.15}
\begin{tabular}{ccccc}
\toprule
$\tau$ & Module & Amp & Prop & Comp \\
\midrule
$0.01$ & $M_2$ &  40 &  5 & 455 \\
       & $M_3$ &   2 &  7 & 491 \\
       & $M_4$ &   0 &  8 & 492 \\
\midrule
$0.05$ & $M_2$ &  40 &  6 & 454 \\
       & $M_3$ &   2 &  7 & 491 \\
       & $M_4$ &   0 &  8 & 492 \\
\midrule
$0.10$ & $M_2$ &  39 &  7 & 454 \\
       & $M_3$ &   2 &  7 & 491 \\
       & $M_4$ &   0 &  8 & 492 \\
\midrule
$0.20$ & $M_2$ &  28 & 63 & 409 \\
       & $M_3$ &   2 & 29 & 469 \\
       & $M_4$ &   0 & 49 & 451 \\
\bottomrule
\end{tabular}
\caption{Per-module fate counts on $\mathcal{D}_{\mathrm{diag}}$
($n_d{=}500$) under varying $\tau$. The compensator-dominant
property reported in \S\ref{ssec:paradox} is robust to the threshold:
$M_3$ has $\geq 469/500$ compensators at every $\tau$ tested, and
the relative ordering of modules by amplifier count
($M_2 \gg M_3 \gg M_4{=}0$) is preserved. $i^\star_{\mathrm{pop}}$
is determined by $\overline{\Delta F_i}$ and is independent of
$\tau$.}
\label{tab:tau-sensitivity}
\end{table}

The reading of the Diagnostic Paradox does not change: $M_3$
remains compensator-dominant at every threshold, $M_2$ remains the
only module with a substantive amplifier population, and the
four-point Pop CCP ordering on $\mathcal{D}_{\mathrm{presc}}$
(App.~\ref{app:cross-agent}) does not depend on $\tau$ because that
comparison uses identical CCP treatment per target and not the
fate label.

\section{$M_1$ CCP Cascade Distributional-Shift Analysis}
\label{app:cascade}

\paragraph{Full results (Table~\ref{tab:linguistic-shift}).}

\begin{table*}[h]
\small\centering
\setlength{\tabcolsep}{6pt}
\renewcommand{\arraystretch}{1.15}
\begin{tabular}{lcc}
\toprule
\textbf{Metric}
  & \textbf{Pop CCP @ $M_3$} & \textbf{Pop CCP @ $M_1$} \\
\midrule
\multicolumn{3}{l}{\emph{Direct shift at patched module (BoW)}} \\
$\cos(M_i^{(0)},\ M_i^{(\mathrm{ccp})})$ & $0.856$ & $0.963$ \\
Categorical disagreement & $9.9\%$ tool-name & surface only \\
\midrule
\multicolumn{3}{l}{\emph{Passive cascade downstream (BoW)}} \\
$\cos(M_2^{(0)},\ M_2^{(\mathrm{ccp}\downarrow)})$ & -- & $0.886$ \\
\ \ intent disagreement & -- & $3.6\%$ \\
$\cos(M_3^{(0)},\ M_3^{(\mathrm{ccp}\downarrow)})$ & -- & $0.928$ \\
\ \ tool-name disagreement & -- & $5.4\%$ \\
$\cos(M_4^{(0)},\ M_4^{(\mathrm{ccp}\downarrow)})$ & $\mathbf{0.837}$ & $\mathbf{0.839}$ \\
\midrule
\multicolumn{3}{l}{\emph{Sentence-encoder (all-MiniLM-L6-v2), per-module}} \\
$\cos(M_1^{(0)},\ M_1^{(\mathrm{ccp/ccp}\downarrow)})$ & $0.984$ & $0.983$ \\
$\cos(M_2^{(0)},\ M_2^{(\mathrm{ccp}\downarrow)})$ & $0.966$ & $0.956$ \\
$\cos(M_3^{(0)},\ M_3^{(\mathrm{ccp/ccp}\downarrow)})$ & $\mathbf{0.920}$ & $\mathbf{0.959}$ \\
$\cos(M_4^{(0)},\ M_4^{(\mathrm{ccp}\downarrow)})$ & $\mathbf{0.906}$ & $\mathbf{0.906}$ \\
\midrule
Net $\Delta F$ & $+0.243^{\ast}$ & $-0.191$ \\
\bottomrule
\end{tabular}
\caption{Cascade distributional-shift comparison on
$\mathcal{D}_{\mathrm{presc}}$ ($n_p{=}111$, baseline $F{=}3.687$).
Top blocks: BoW cosine as a conservative lower bound. Bottom
block: sentence-encoder (all-MiniLM-L6-v2) cosine, computed on
the same per-module outputs (\S\ref{ssec:mechanism}); the
$M_4$ magnitude invariance (both $0.906$) holds under both
metrics (paired Wilcoxon $p{=}0.76$), and the $M_3$ direct
shift is significantly larger under $M_3$ CCP than under $M_1$
CCP ($0.920$ vs.\ $0.959$, $p{=}0.006$). This rules out the
concern that BoW is too coarse to detect the cascade. Dominant
failure mode under Pop CCP @ $M_3$: slot-value format error.}
\label{tab:linguistic-shift}
\end{table*}

\paragraph{Motivation.}
Section~\ref{ssec:mechanism} argues that the Diagnostic
Paradox is governed by the \emph{direction} of downstream
perturbation, not its magnitude. This appendix provides the full
methodology and results.

\paragraph{Procedure.}
We run the Pop CCP at $M_1$ configuration ($k{=}5$ severity-ranked
corrections, same judge and correction-selection rule as the Pop
CCP at $M_3$ configuration) on $\mathcal{D}_{\mathrm{presc}}$
($n_p{=}111$, baseline $\bar F{=}3.687$). For each task and each
module $j \in \{M_1, M_2, M_3, M_4\}$, we record the raw text
output and compare it to the baseline using:
(i)~BoW cosine similarity; (ii)~categorical disagreement at $M_2$
(intent label) and $M_3$ (tool\_name); (iii)~slot-keys Jaccard
at $M_2$ to separate schema-level from value-level shift.

\paragraph{Hypotheses stated prior to running this configuration.}
\begin{itemize}\itemsep=2pt
\item \textbf{Hypothesis A} (strict contract-preservation):
  $\cos(M_2^{(0)}, M_2^{(\mathrm{ccp}\downarrow)}) > 0.95$;
  Pop CCP at $M_1$ hands downstream a cleaner input without
  disturbing inter-module agreement.
  This was not supported: measured $M_2$ cosine was $0.886$.
\item \textbf{Hypothesis B} (directional account):
  Pop CCP at $M_1$ induces downstream shift of comparable
  magnitude to Pop CCP at $M_3$, but along a quality-aligned
  direction, so net quality gain exceeds disruption cost.
  This was consistent with the data: terminal $M_4$ cosine
  $0.839$ (Pop CCP @ $M_1$) $\approx$ $0.837$ (Pop CCP @ $M_3$),
  yet $\Delta F{=}{-}0.191$ (improvement) vs.\ $+0.243$ (Hazard).
\end{itemize}

\section{Inter-Judge Agreement Validation}
\label{app:judge-agreement}

To validate Gemini~2.5~Flash as the judge, we re-scored a
stratified sample of $50$ $\mathcal{D}_{\mathrm{diag}}$ tasks
across all four modules ($n{=}200$ assessments) using GPT-4o
as an independent cross-family judge with the same rubric
(\S\ref{ssec:fi}).

\paragraph{Results.}
Krippendorff's $\alpha{=}0.772$ and Pearson $r{=}0.814$
($p{<}0.001$; Spearman $r{=}0.782$) indicate substantial
agreement. Mean absolute severity difference is $0.088$
(on a $[0,0.99]$ scale). Per-module: $M_1$ ($\alpha{=}0.657$),
$M_2$ ($\alpha{=}0.686$), $M_3$ ($\alpha{=}0.128$),
$M_4$ ($\alpha{=}0.717$).

$M_3$'s low $\alpha$ reflects ceiling compression (most tasks
score $\approx0.95$ under both judges). Three complementary
$M_3$-specific checks confirm the core finding is not a
judge artifact: raw Pearson $r{=}0.615$ ($p{<}0.001$) shows
substantial linear correlation; binary agreement at
$\mathrm{sev}{\geq}0.5$ is $80\%$; and the $M_3{>}M_1$
direction is preserved on $42/50$ tasks ($84\%$) under GPT-4o,
ruling out a single-judge artifact for the Diagnostic Paradox.

\section{$F$ Validity Against Behavioral Tool-Match}
\label{app:f-validity}

A common concern with LLM-judge metrics is that $F$ might be
internally consistent yet disconnected from the agent's actual
behavior. We test this by correlating per-task $F$ with the
\textbf{first-action tool-name match}: a deterministic, non-judge
behavioral metric that compares the agent's first emitted tool
name against the cached oracle gold (no judge involvement).
Results are in Table~\ref{tab:f-validity}.

\begin{table*}[h]
\centering\small
\setlength{\tabcolsep}{10pt}
\renewcommand{\arraystretch}{1.15}
\begin{tabular}{@{}l r r r r@{}}
\toprule
\textbf{Split} & $n$ & \textbf{Pearson $r$} & \textbf{Spearman $\rho$} & $\Delta\bar{F}_{\mathrm{miss{-}match}}$ \\
\midrule
$\mathcal{D}_{\mathrm{presc}}$ baseline (gpt-4o-mini) & 111 & $-0.902^{\ast\ast\ast}$ & $-0.829^{\ast\ast\ast}$ & $+2.81$ \\
$\mathcal{D}_{\mathrm{presc}}$ all configs (gpt-4o-mini) & 111 & $-0.867^{\ast\ast\ast}$ & $-0.850^{\ast\ast\ast}$ & $+2.70$ \\
$\mathcal{D}_{\mathrm{diag}}$ (gpt-4o-mini)  & 500 & $-0.453^{\ast\ast\ast}$ & $-0.262^{\ast\ast\ast}$ & $+2.43$ \\
$\mathcal{D}_{\mathrm{diag}}$ (Llama~4 Scout) & 500 & $-0.357^{\ast\ast\ast}$ & $-0.250^{\ast\ast\ast}$ & $+1.50$ \\
\bottomrule
\end{tabular}
\caption{$F$ correlates strongly with first-action tool-name match across splits and agent families.
$^{\ast\ast\ast}p{<}0.001$.
$\Delta\bar{F}_{\mathrm{miss{-}match}}$ = mean $F$ on tool-match-failure tasks minus mean $F$ on
tool-match-success tasks. Qwen3-32b is excluded as the $\mathcal{D}_{\mathrm{diag}}$ sample
($n{=}50$) is too small for reliable correlation estimation.}
\label{tab:f-validity}
\end{table*}

On the prescription split where the Diagnostic Paradox lives,
$F$ correlates with behavioral tool-match at $r{=}{-}0.90$
($p{<}10^{-40}$), with tasks failing tool-name selection scoring
$2.81$ $F$-units higher on average than tasks succeeding. The
correlation is weaker but still highly significant on
$\mathcal{D}_{\mathrm{diag}}$, where tool-match ceiling effects
compress the range ($97.6\%$ match rate for gpt-4o-mini means only
$n{=}12$ failure points). This rules out the concern that the
$M_1$-vs.-$M_3$ Paradox is an artifact of $F$ being judge-internal:
when the agent fails to call the gold tool, $F$ rises sharply,
and the directional relationship holds across both cross-family
agents.

\section{Linguistic Typology of \texorpdfstring{$M_3$}{M3} Routing Errors}
\label{app:linguistic-typology}

App.~\ref{app:cross-agent} shows that oracle injection
at $M_3$ yields the largest improvement of any configuration
(gpt-4o-mini $\Delta{=}{-}2.007$; Llama~4 Scout $\Delta{=}{-}1.795$),
confirming that $M_3$ is improvable when given the exact gold
output---yet CCP at $M_3$ causes harm.
A coarse classification of $M_3$ baseline-vs.-oracle differences
on a subsample of $\mathcal{D}_{\mathrm{presc}}$ explains why:

\begin{itemize}\itemsep=1pt
\item \emph{Slot-value format error} (majority):
  $M_3$ selects the correct tool but provides arguments in the
  wrong format (e.g., a descriptive item name instead of a
  catalog numeric ID). Oracle injection directly substitutes
  the gold \texttt{tool\_args}.
\item \emph{Tool-name mismatch}: $M_3$ selects an
  adjacent-but-wrong tool (e.g., \texttt{modify\_pending\_order\_items}
  vs.\ \texttt{modify\_pending\_order\_address}).
\item \emph{Argument sigil variance}: minor string-format
  differences in order IDs or payment method identifiers.
\end{itemize}

Category (i) dominates. When CCP injects correction triples of
the form (input, wrong \texttt{tool\_args}, oracle \texttt{tool\_args}),
it shifts $M_3$'s output distribution toward oracle argument
formats---an executable-semantic change that downstream $M_4$,
co-adapted to the original noisy distribution, cannot absorb.
This is the Linguistic Contract at work: the oracle injection
upper bound is achievable, but only because it substitutes the
output directly; CCP, which shifts the prompt, instead disrupts
the downstream alignment and produces the Hazard.

\end{document}